\documentclass[11pt]{article}

\usepackage[preprint]{acl}

\usepackage{times}
\usepackage{latexsym}

\usepackage[T1]{fontenc}

\usepackage[utf8]{inputenc}

\usepackage{microtype}

\usepackage{inconsolata}

\usepackage{graphicx}
\usepackage{booktabs}
\usepackage{makecell}
\usepackage{url}
\usepackage{amsmath}
\usepackage[table]{xcolor}
\usepackage{multirow} 
\usepackage{tabularx}
\usepackage{float}
\usepackage{placeins}
\usepackage{fontawesome5}
\usepackage{makecell}
\usepackage{array}
\newcolumntype{C}[1]{>{\centering\arraybackslash}p{#1}}

\newcommand{\taxonomy}[0]{\textsc{LiTEx}}
\definecolor{mycolor}{RGB}{217, 217, 217}

\title{Agree, Disagree, Explain: Decomposing Human Label Variation \\ in NLI through the Lens of Explanations}

\author{
\textbf{Pingjun Hong\thanks{\; Equal contribution.}\textsuperscript{\faLaptopCode\kern1pt\faGraduationCap}} \quad
\textbf{Beiduo Chen\footnotemark[1]\textsuperscript{\faMountain\kern1pt\faRobot}} \quad
\textbf{Siyao Peng\textsuperscript{\faMountain\kern1pt\faRobot}} \quad
\\
\textbf{Marie-Catherine de Marneffe\textsuperscript{\faPenFancy}} \quad
\textbf{Benjamin Roth\textsuperscript{\faLaptopCode\kern1pt\faBook}} \quad
\textbf{Barbara Plank\textsuperscript{\faMountain\kern1pt\faRobot}}
\\[8pt]
\textsuperscript{\faLaptopCode}Faculty of Computer Science, \textsuperscript{\faGraduationCap}UniVie Doctoral School Computer Science, \\\textsuperscript{\faBook} Faculty of Philological and Cultural Studies, University of Vienna, Austria \\
\textsuperscript{\faMountain}MaiNLP, Center for Information and Language Processing, LMU Munich, Germany \\
\textsuperscript{\faRobot}Munich Center for Machine Learning, Germany
\textsuperscript{\faPenFancy}FNRS, CENTAL, UCLouvain, Belgium 
\\[6pt]
{
\footnotesize
{\tt{
\{\href{mailto:pingjun.hong@univie.ac.at}{\textcolor{black}{pingjun.hong}},
\href{mailto:benjamin.roth@univie.ac.at}{\textcolor{black}{benjamin.roth}}\}@univie.ac.at,
\{\href{mailto:beiduo.chen@lmu.de}{\textcolor{black}{beiduo.chen}},
\href{mailto:b.plank@lmu.de}{\textcolor{black}{b.plank}}\}@lmu.de}},}
\\ 
{\footnotesize {\tt{ \href{mailto:loganpeng1992@gmail.com}{\textcolor{black}{loganpeng1992@gmail.com}},
\href{mailto:marie-catherine.demarneffe@uclouvain.be}{\textcolor{black}{marie-catherine.demarneffe@uclouvain.be}}}}}
}

\begin{document}

\maketitle

\begin{abstract}

Natural Language Inference (NLI) datasets often exhibit human label variation. To better understand these variations, explanation-based approaches analyze the underlying reasoning behind annotators' decisions. One such approach is the \taxonomy{} taxonomy, which categorizes free-text explanations in English into reasoning categories. However, previous work applying \taxonomy{} has focused on within-label variation: cases where annotators agree on the NLI label but provide different explanations. This paper broadens the scope by examining how annotators may diverge not only in the reasoning category but also in the labeling. We use explanations as a lens to analyze variation in NLI annotations and to examine individual differences in reasoning.
We apply \taxonomy{} to two NLI datasets and align annotation variation from multiple aspects: NLI label agreement, explanation similarity, and taxonomy agreement, with an additional compounding factor of annotators' selection bias. 
We observe instances where annotators disagree on the label but provide similar explanations, suggesting that surface-level disagreement may mask underlying agreement in interpretation. Moreover, our analysis reveals individual preferences in explanation strategies and label choices. These findings highlight that agreement in reasoning categories better reflects the semantic similarity of explanations than label agreement alone.
Our findings underscore the richness of reasoning-based explanations and the need for caution in treating labels as ground truth.

\end{abstract}

\section{Introduction}\label{sec:introduction}
Natural Language Inference (NLI) has long served as a benchmark for language understanding \citep{inproceedings, condoravdi-etal-2003-entailment}, 
yet annotation divergence, both in labels and in underlying reasoning, has been recognized as a challenge. 
Recent works have provided new explanations of this phenomenon, drawing on insights from linguistics, pragmatics, and conceptual framing \citep{plank-2022-problem, jiang-etal-2023-ecologically, kalouli-etal-2023-curing, gubelmann-etal-2023-truth}.

\begin{figure}[t]
  \includegraphics[width=\columnwidth]{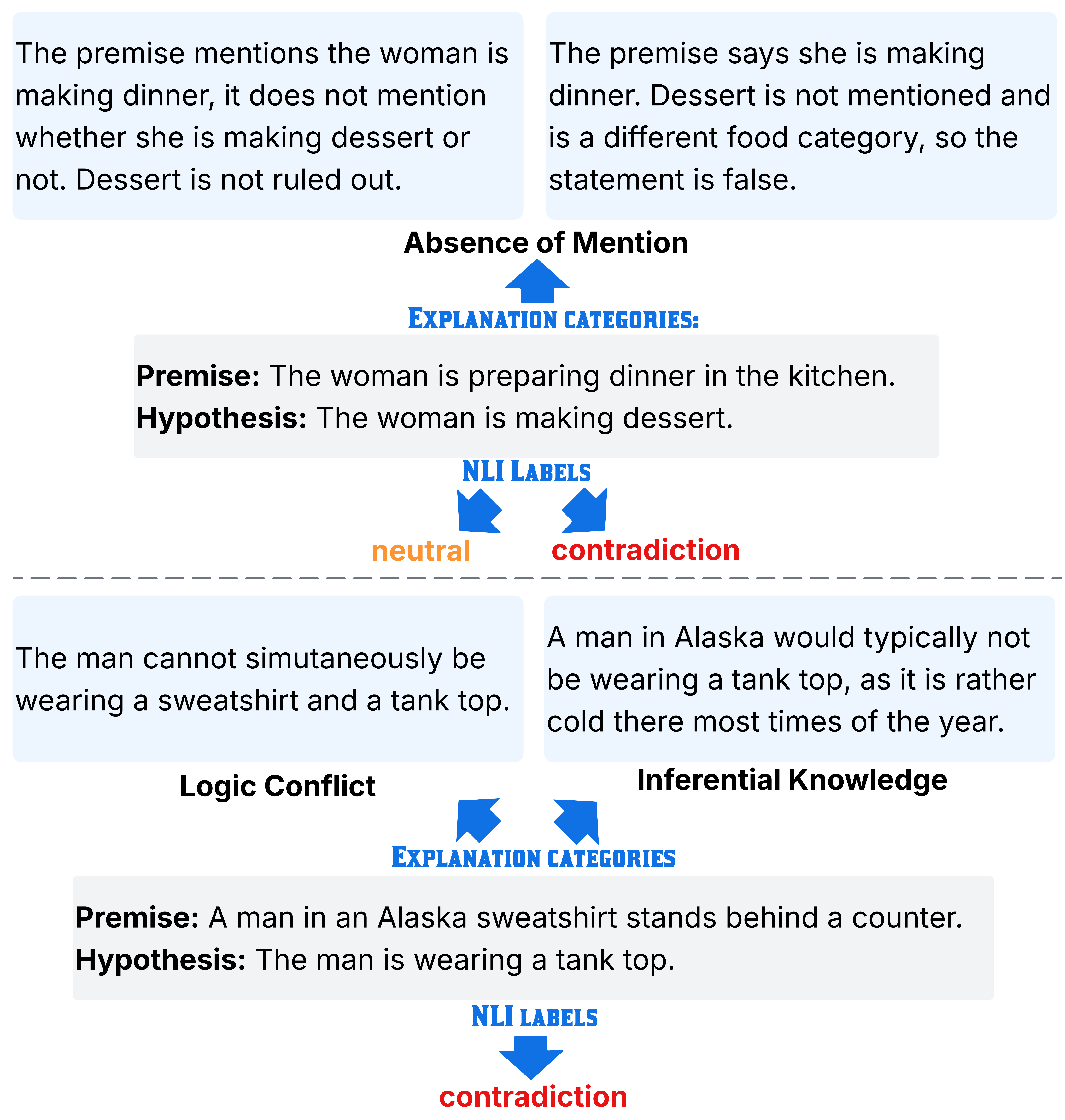}
  \caption{\textbf{Agreement and disagreement in NLI annotations through labels and explanations.} The two examples illustrate how annotators may diverge in NLI labels or explanation categories, featuring the \taxonomy{} categories \textit{Logical Conflict} and \textit{Absence of Mention}.
}
  \label{fig:intro}
\end{figure}

Explanations written in free-text provide valuable insights into the reasons behind label assignment \citep{jiang-etal-2023-ecologically, tan-2022-diversity,chen-etal-2025-threading}. Instead of treating NLI labels as isolated outcomes, explanations reveal underlying reasoning strategies that annotators employ. Building on \citet{jiang-marneffe-2022-investigating}, which focuses on categorizing linguistic sources of disagreement in the premise–hypothesis pair, 
\citet{hong2025litexlinguistictaxonomyexplanations} introduced the  \taxonomy{} taxonomy. The taxonomy is developed based on the e-SNLI dataset \citet{camburu2018esnlinaturallanguageinference} and categorizes explanations according to eight distinct reasoning strategies.

The \taxonomy{} taxonomy provides a structured framework for analyzing free-text explanations jointly with NLI labels. By considering both label assignments and explanation categories, we can uncover patterns of agreement and disagreement. These distinctions would remain hidden if one only considered the NLI label distribution alone. Incorporating explanation-based reasoning further helps illuminate potential sources of variation in NLI.

Figure~\ref{fig:intro} illustrates two representative cases of agreement and disagreement when looking into NLI labels and explanations using the \taxonomy{} categories.
The top example shows a scenario in which both annotators rely on the \textit{Absence of Mention} rationale, yet assign different labels (neutral vs.\ contradiction), highlighting divergence in label judgments despite similar explanation strategies. The bottom example shows annotators agreeing on the contradiction label while providing explanations grounded in different categories, reflecting variation in how the relationship between the premise and hypothesis is interpreted.

While \taxonomy{} was originally designed to study within-label variation (like the second example in Figure~\ref{fig:intro}, which provides different explanations for the same NLI label), it remains unclear how well its approach generalizes to label variation settings. In this work, we analyze the interaction of label choices, explanation categories, and explanation text similarities to provide a deeper view of how explanations relate to NLI labels. This perspective allows us to identify patterns of agreement and disagreement that are not apparent from label distribution alone and to better characterize sources of variation across annotators. Specifically, our key contributions include:

\paragraph{From within-label to label variation} 
We annotate two NLI datasets, LiveNLI \citep{jiang-etal-2023-ecologically} and VariErr \citep{weber-genzel-etal-2024-varierr} with the \taxonomy{} taxonomy, enabling the study of reasoning categories across both within-label and label variation. 

\paragraph{Annotator tracking via explanation categories} 
By combining NLI labels with explanation categories, we track individual annotators' reasoning patterns and uncover behavioral consistencies that are not apparent from label distributions alone.

\paragraph{Quantitative analysis of agreement beyond labels} 
We measure agreement at three levels: NLI label, explanation category, and text similarity of explanations. Our results show that alignment in reasoning categories better correlates with explanation similarity than NLI label agreement alone, emphasizing the importance of explanations for understanding annotator disagreement.

\section{Related Work}\label{sec:related-work}

Most benchmark NLI datasets provide multiple annotations per instance, 
enabling the study of annotation variation. 
For example, the Stanford NLI (SNLI; \citealt{bowman-etal-2015-large}) and MultiNLI \citep{williams-etal-2018-broad} datasets 
collect multiple crowd-sourced judgments for each premise–hypothesis pair, 
motivating work on systematic disagreement and label variation, often through re-annotation and analyses specifically examining disagreement \citep{pavlick-kwiatkowski-2019-inherent, kalouli-etal-2023-curing}. To further address this, adversarial datasets such as ANLI \citep{nie-etal-2020-adversarial} and resources targeting different aspects of annotation variation, such as ChaosNLI \citep{nie-etal-2020-learn} for annotator disagreement and AmbiEnt \citep{liu-etal-2023-afraid} for ambiguity, are introduced. 

Closer to our line of work, recent datasets use free-text explanations and highlights to reveal the reasoning behind NLI labels \citep{camburu2018esnlinaturallanguageinference, jiang-etal-2023-ecologically, nighojkar-etal-2023-strong, weber-genzel-etal-2024-varierr}. 
\textbf{e-SNLI} expanded SNLI \citep{bowman-etal-2015-large} by crowd-sourcing highlight and explanation annotations on the pre-annotated majority label.
\textbf{LiveNLI} \citep{jiang-etal-2023-ecologically} recruited crowd-workers to annotate NLI labels while also providing highlights and explanations (i.e., ecologically valid explanations, produced jointly with the label rather than post hoc).
\textbf{VariErr} \citep{weber-genzel-etal-2024-varierr} use such ecologically valid explanations as a foundation for error detection.
\citet{hong2025litexlinguistictaxonomyexplanations} built a taxonomy, \textbf{\taxonomy{}}, to categorize these free-text explanations, but its scope was limited to e-SNLI and focused on analyzing within-label variation.

A parallel line of work has examined \textit{annotator disagreement}. For example, \citet{demarneffe-etal-2012} and \citet{Uma2022ScalingAD} identified structured patterns of disagreement. For NLI, understanding human reasoning is crucial to interpreting agreement or disagreement.
Annotators often rely on various reasoning strategies, such as substitution, negation, bridging inferences, and world knowledge inference \citep{jiang-marneffe-2022-investigating,kalouli-etal-2023-curing, sanyal-etal-2024-machines,hong2025litexlinguistictaxonomyexplanations}. 

Prior work on annotator decisions often focuses on the relationship between the premise-hypothesis pair and the resulting NLI label, but the corelation with explanations remains underexplored. In particular, it is unclear how similar explanation categories can lead to different labels, or how the same label may reflect distinct rationales. We address this gap by using the \taxonomy{} taxonomy to analyze the interaction between explanation categories and label assignments, providing a more fine-grained view of variations in NLI annotations.

\section{From Within-label Variation to Label Variation}\label{sec:applicability}

To characterize patterns of variation in different NLI datasets, we use the \taxonomy{} taxonomy \citep{hong2025litexlinguistictaxonomyexplanations} and apply it to two additional English NLI datasets exhibiting label variation. 
We then examine the co-occurrence patterns 
between NLI labels and explanation categories across datasets.

\subsection{\taxonomy{}: a Linguistic Taxonomy of Explanations}
\taxonomy{} categorizes NLI explanations into two reasoning categories: \textit{Text-Based (TB)} and \textit{World-Knowledge (WK)}. 
TB draws on linguistic evidence in the premise and hypothesis and comprises six categories: \textit{Coreference}, \textit{Syntactic}, \textit{Semantic}, \textit{Pragmatic}, \textit{Absence of Mention}, and \textit{Logic Conflict}. 
WK invokes background knowledge beyond text, covering \textit{Factual Knowledge} and \textit{Inferential Knowledge}. 
For detailed definitions and annotated examples of all categories, please refer to \citet{hong2025litexlinguistictaxonomyexplanations}.

\taxonomy{} was originally developed on free-text explanations in e-SNLI \citep{camburu2018esnlinaturallanguageinference} to characterize within-label variation, where annotators reach the same label via different rationales \citep{jiang-etal-2023-ecologically, hong2025litexlinguistictaxonomyexplanations}. 
In this paper, we use \taxonomy{} beyond its original scope by applying it to additional datasets, examining reasoning strategies not only within-label but also across labels. Furthermore, we use the taxonomy to analyze annotator labeling behavior and to deepen our understanding of the relationship between explanations and NLI labels.

\subsection{Annotation on LiveNLI and VariErr}

\begin{figure}[t]
  \includegraphics[width=\columnwidth]{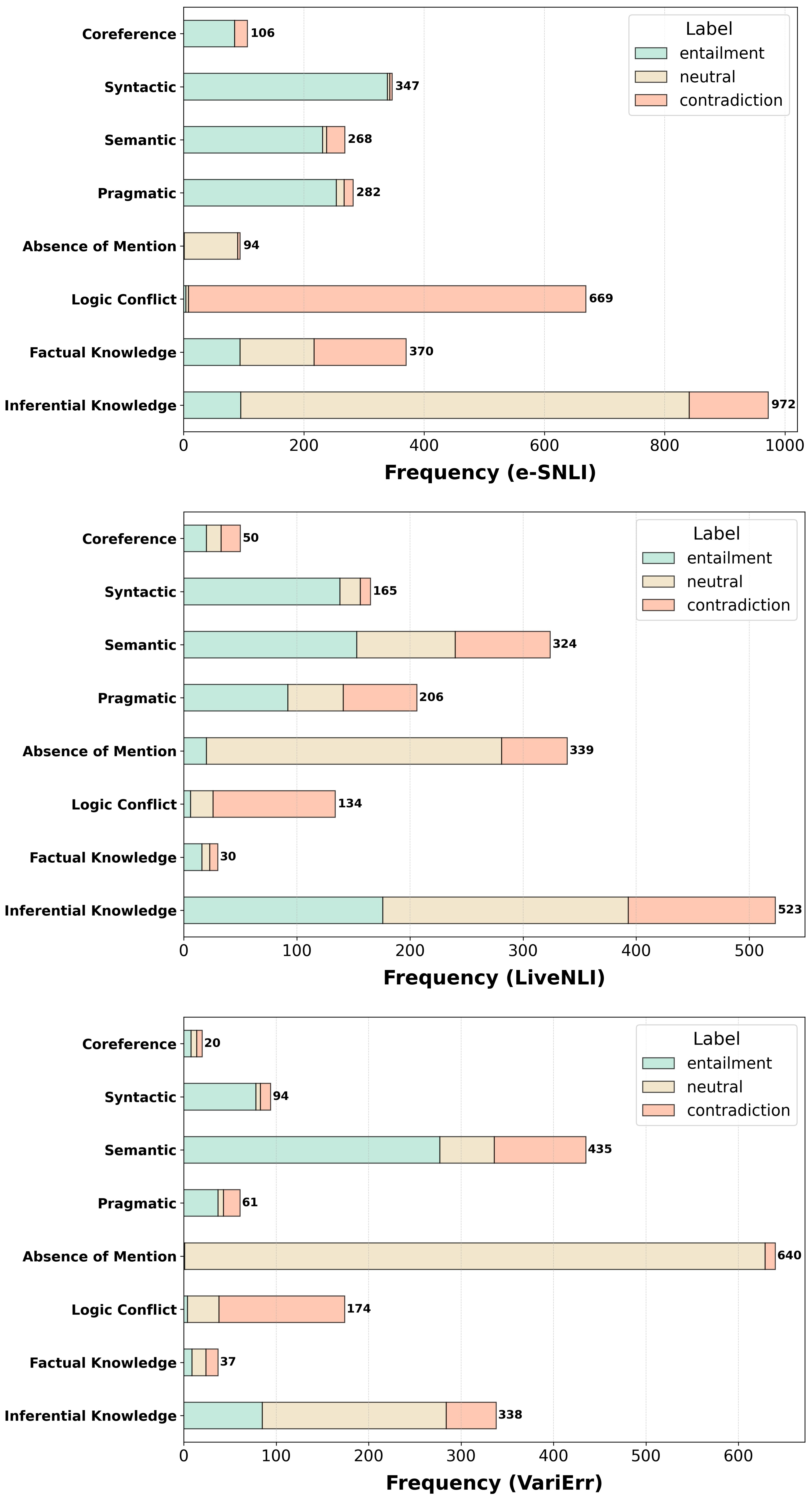}
  \caption{Co-occurrence of \taxonomy{} explanation categories and NLI labels across the e-SNLI, LiveNLI, and VariErr datasets.}
  \label{fig:cooccurrence}
\end{figure}

To study how annotators diverge in both reasoning and label selection, we apply \taxonomy{} to two English datasets with label variation and explanation annotations — \textbf{LiveNLI} \citep{jiang-etal-2023-ecologically} and \textbf{VariErr} \citep{weber-genzel-etal-2024-varierr}.

LiveNLI is a high-quality explanation dataset derived from a subset of MNLI \citep{williams-etal-2018-broad}, containing 122 NLI items. Each item is annotated by at least 10 crowdworkers, who assign one or more NLI labels (\textit{true}, \textit{either}, \textit{false}), highlight relevant spans, and provide free-text explanations \citep{jiang-etal-2023-ecologically}. The dataset contains 1,404 explanation-label pairs. For the consistency of our analyses and to align with e-SNLI and VariErr, we map (\textit{true}, \textit{either}, \textit{false}) to (\textit{entailment}, \textit{neutral}, \textit{contradiction}).

VariErr complements LiveNLI by focusing on variation and errors in English NLI. It consists of 1,933 model-generated explanations for 500 re-annotated MNLI items, along with 7,732 human validity judgments \citep{weber-genzel-etal-2024-varierr}. LiveNLI captures natural annotator disagreement, and VariErr introduces both plausible alternative explanations and annotation errors.

A key difference from e-SNLI lies in the labeling scheme. While e-SNLI assigns a single gold label, both LiveNLI and VariErr allow multiple plausible labels per instance and are ecologically valid, with labels and explanations annotated by the same people. These features allow us to examine the relationship between label assignments and explanation categories. Appendix~\ref{sec:label_distribution} presents a detailed analysis of label distribution per NLI item in the two datasets. 

We use \taxonomy{} to annotate all explanations from LiveNLI and VariErr.
All annotations are carried out by a trained annotator. The annotator is instructed to categorize free-text explanations according to the reasoning explicitly expressed within them and is asked to select the most prominent explanation categories from \taxonomy{}. This design follows the original setup in which \taxonomy{} was introduced, ensuring alignment with prior work.
To measure inter-annotator agreement (IAA), we recruited a second annotator to annotate 100 explanations from each dataset independently.\footnote{Both annotators were students trained with the taxonomy definitions and examples, and were paid according to the national standard.} 
We obtained a Cohen's Kappa ($\kappa$) of 0.828 for LiveNLI and 0.792 for VariErr, similar to the IAA on e-SNLI \citep{hong2025litexlinguistictaxonomyexplanations}. For a more detailed analysis of IAA results and per-category agreement, please refer to Appendix~\ref{sec:IAA}.

\subsection{\taxonomy{} Categories across NLI Labels}

Figure~\ref{fig:cooccurrence} shows the distributions of the \taxonomy{} categories across NLI labels in three datasets: e-SNLI, LiveNLI and VariErr.
In terms of frequency of the \taxonomy{} categories, we see that \textit{Coreference} appears less frequently in all datasets compared to the other categories. Notable differences can also be observed across datasets. For example, \textit{Inferential Knowledge} is the dominant category in both e-SNLI and LiveNLI but is less prominent in VariErr. In contrast, \textit{Absence of Mention} is the most frequent category in VariErr, ranks second in LiveNLI, and occurs relatively less often in e-SNLI.

Additional interesting patterns arise in the \textbf{co-occurrences} between the distributions of NLI labels within taxonomy categories. The \textit{neutral} label dominates the \textit{Absence of Mention} category across all three datasets. This is consistent with the nature of this reasoning category, which focuses on information gaps between the premise and hypothesis. For \textit{Factual Knowledge} and \textit{Inferential Knowledge}, explanations are distributed relatively evenly across the three NLI labels. This reflects the fact that these categories involve the introduction of external or inferred knowledge, without a strong bias toward a specific label. 
The \textit{Syntactic}, \textit{Semantic}, and \textit{Pragmatic} categories are more strongly associated with entailment (despite slight variations across datasets), suggesting that annotators often rely on evidence from different linguistic levels within the premise and hypothesis to establish entailment relationships.

In sum, although the absolute distributions of explanation categories differ across datasets, the label distribution and dominant label for each category remain highly consistent. The observed differences in category distribution may stem from factors such as annotator backgrounds and preferences, dataset-specific item selection, and annotation guidelines.  
Nevertheless, the stable category–label co-occurrence patterns indicate that, , despite being originally developed only on the e-SNLI dataset, \taxonomy{} provides \textbf{a reliable characterization of reasoning categories that generalize across explanations} in different NLI datasets.

\section{Label and Reasoning Preferences among Individual Annotators}
\label{sec:label-category-distribution}

To better understand disagreement in NLI, we analyze annotator preferences over NLI labels and explanation categories. We focus on four annotators in LiveNLI and VariErr, as e-SNLI does not provide annotator identifiers for tracking.
Specifically, we track individual annotators across two dimensions: their \textbf{NLI label preferences} (e.g., tendency to overuse ``neutral"), and their \textbf{reasoning-category preferences},taxonomy classification of their free-text explanations annotated in this paper.

\begin{figure}[t]
  \includegraphics[width=\columnwidth]{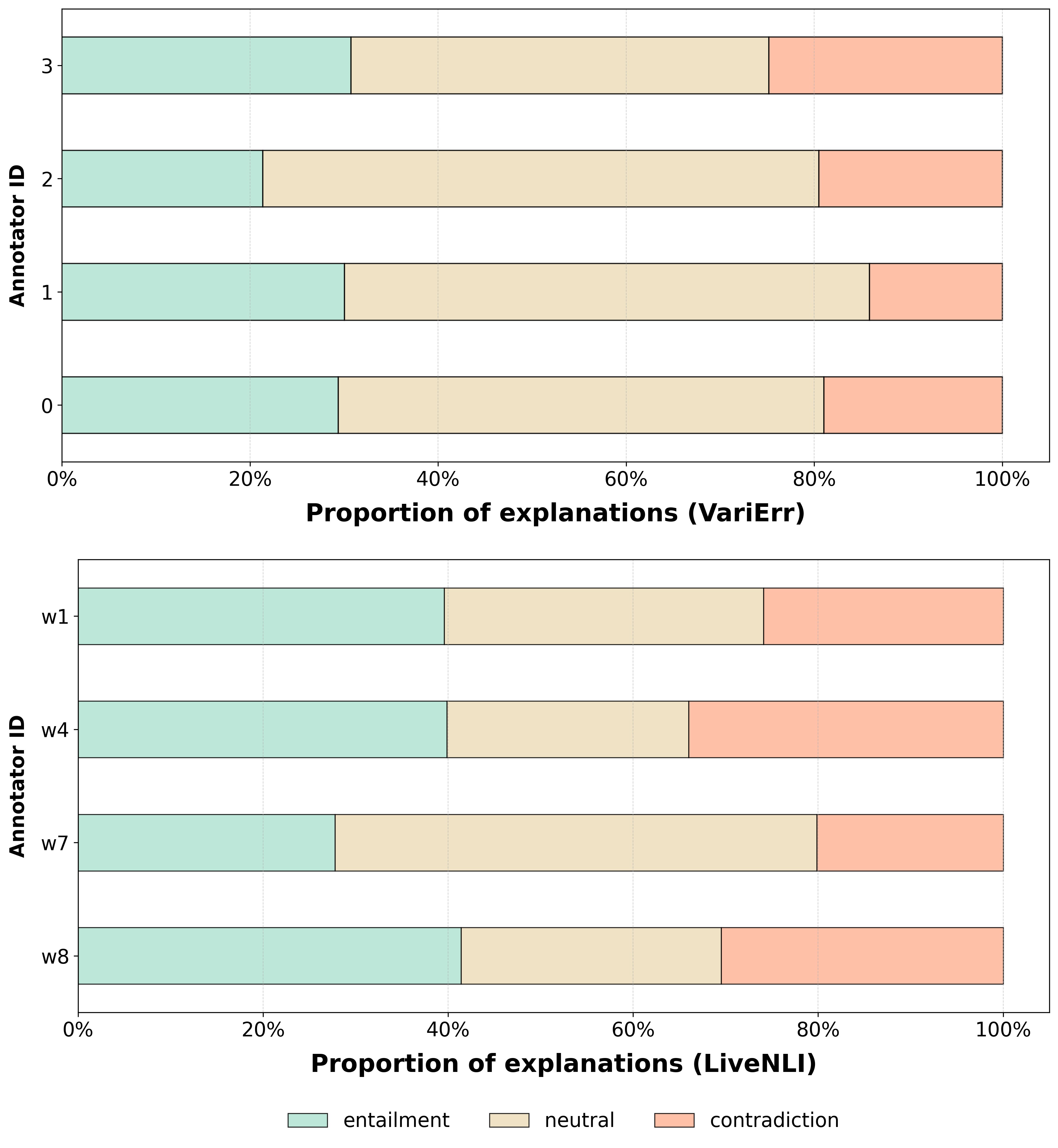}
  \caption{Distribution of NLI labels (entailment, neutral, contradiction) across LiveNLI and VariErr annotators. The legend at the bottom specifies the color–label correspondence, while the area of each color segment represents the number of instances assigned to that label.}
  \label{fig:nlilabel_varierr}
\end{figure}

\begin{figure}[t]
    \centering
    \includegraphics[width=\columnwidth]{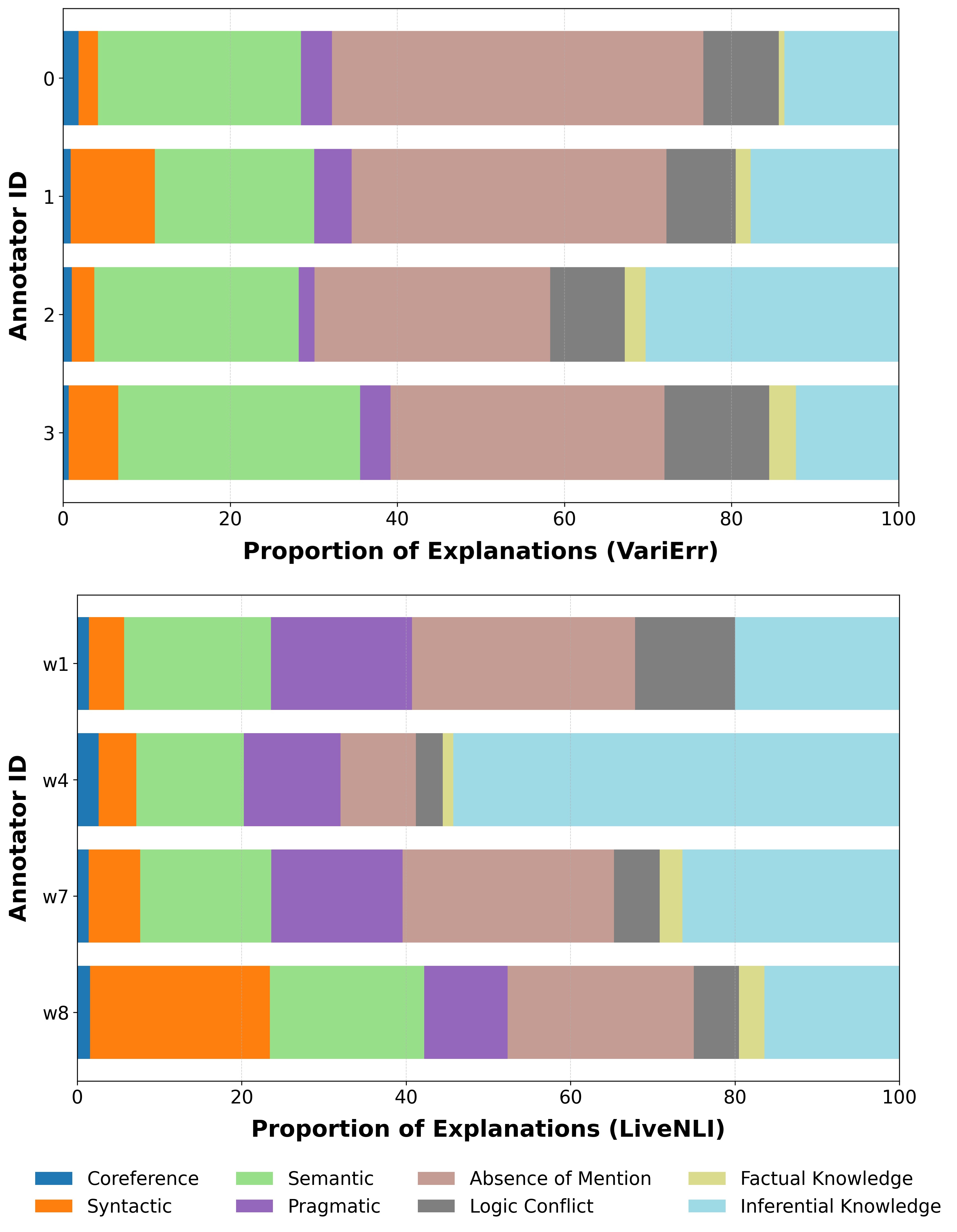}
    \caption{Distribution of explanation category per annotators in LiveNLI and VariErr. Colors correspond to different explanation categories.}
    \label{fig:annotator_taxonomy}
\end{figure}

Unlike VariErr, in which all items are annotated by the same four annotators, LiveNLI involves a much larger pool of annotators \citep{jiang-etal-2023-ecologically}. The number of items annotated by the same annotator ranges from 1 to 122. To align with the setup in VariErr, we chose a group of four annotators \{\texttt{w1}, \texttt{w4}, \texttt{w7}, \texttt{w8}\} from LiveNLI who have the highest number of overlapping annotated items (115 in total). 
Appendix~\ref{sec:distribution_counts} provides detailed tables of NLI label distributions and explanation category preferences for the four annotators in LiveNLI and the four annotators of VariErr, computed over the intersection of items they annotated.

\paragraph{Label Preferences}
Figure~\ref{fig:nlilabel_varierr} illustrates the distribution of NLI labels assigned by four annotators for VariErr and for LiveNLI. For VariErr, while all annotators exhibit a noticeable preference for the \textit{neutral} label, the degree of this preference varies. Annotators \texttt{0}, \texttt{1}, and \texttt{2} all assign \textit{neutral} in over 50\% of their examples, with Annotator \texttt{2} showing the strongest bias—nearly 60\% of their annotations are \textit{neutral}. In contrast, Annotator \texttt{3} demonstrates a more balanced labeling behavior, with a more even spread across \textit{entailment}, \textit{neutral}, and \textit{contradiction}, resulting in the lowest \textit{neutral} proportion (44.42\%).

As for LiveNLI, three annotators (\texttt{w1}, \texttt{w4}, \texttt{w8}) show a slight preference for the \textit{entailment} label, each assigning it to approximately 40\% of the items. In contrast, annotator \texttt{w7} exhibits a stronger preference for the \textit{neutral} label, assigning it in over half of the cases (52.08\%). The \textit{contradiction} labels are the least common overall. Compared to annotators in VariErr, where \textit{neutral} was the dominant label for all four annotators, the LiveNLI group shows more varied labeling tendencies.

\begin{table*}[t!hb]
\centering
\resizebox{\textwidth}{!}{%
\begin{tabular}{lrr|r|rrrrrr}
\toprule
\begin{tabular}[c]{@{}c@{}}\textbf{Agreement}\\\textbf{Class}\end{tabular} & \textbf{Entropy} 
& \begin{tabular}[c]{@{}c@{}}\textbf{Support}\\\textbf{(\%)}\end{tabular}
& \begin{tabular}[c]{@{}c@{}}\textbf{Category}\\ \textbf{Agreement}\end{tabular}
& \begin{tabular}[c]{@{}c@{}}\textbf{token}\\\textbf{1-gram}\end{tabular} & \begin{tabular}[c]{@{}c@{}}\textbf{token}\\\textbf{2-gram}\end{tabular} & \begin{tabular}[c]{@{}c@{}}\textbf{POS}\\\textbf{1-gram}\end{tabular} & \begin{tabular}[c]{@{}c@{}}\textbf{POS}\\\textbf{2-gram}\end{tabular} 
& \begin{tabular}[c]{@{}c@{}}\textbf{cosine}\\\textbf{(\%)}\end{tabular} & \begin{tabular}[c]{@{}c@{}}\textbf{euclidean}\\\textbf{(\%)}\end{tabular} \\
\midrule
\rowcolor{mycolor}\multicolumn{10}{l}{\textit{VariErr}\rule{0pt}{8pt}} \\
Full (4-0-0)  & 0.00     & 43.75  & \textbf{0.76}    & \textbf{35.05}   & \textbf{11.53}    &  \cellcolor{purple!20} 74.21    & \textbf{35.06}    & \textbf{52.87} &  \cellcolor{purple!20} 51.89  \\
Partial (3-1-0) & 0.81  & 28.95 & 0.60   & 34.72   & 10.62   &  \cellcolor{cyan!20} \textbf{78.31}    & 34.85   & 52.81 &  \cellcolor{cyan!20}  \textbf{51.96}  \\
Two Pairs (2-2-0) & 1.00 & 23.36 & 0.56    &  \cellcolor{purple!20} 30.80    &  \cellcolor{purple!20} 8.50   &   73.47   &  \cellcolor{purple!20} 31.23   & 49.22 & 51.02  \\
Divergent (2-1-1) & 1.50 & 3.95  & 0.50  &  \cellcolor{cyan!20} 32.02   &  \cellcolor{cyan!20} 9.96   & 70.37    &  \cellcolor{cyan!20} 31.50   & 48.21 & 50.91   \\
\rowcolor{mycolor}\multicolumn{10}{l}{\textit{LiveNLI}\rule{0pt}{8pt}} \\
Full (4-0-0)   & 0.00     & 21.74 &  \textbf{0.62} &  \cellcolor{purple!20} 40.31 &  \cellcolor{purple!60} 10.26 & \textbf{88.89} &  \cellcolor{purple!60} 41.96 & \textbf{58.05} & \textbf{53.24}  \\
Partial (3-1-0)   & 0.81  & 34.78 & \cellcolor{purple!20} 0.56 &  \cellcolor{cyan!20} \textbf{40.44} &  \cellcolor{cyan!20} \textbf{11.95} &   \cellcolor{purple!20} 86.99 &  \cellcolor{cyan!20} \textbf{44.02} &  \cellcolor{purple!20} 54.47 &  \cellcolor{purple!60} 52.27  \\
Two Pairs (2-2-0) & 1.00  & 23.48 & \cellcolor{cyan!20} 0.60  & 38.97 &  \cellcolor{cyan!20} 10.67 &  \cellcolor{cyan!20} 88.38 &  \cellcolor{cyan!20} 43.24 & \cellcolor{cyan!20} 55.33 &  \cellcolor{cyan!20} 52.84 \\
Divergent (2-1-1) & 1.50 & 20.00 & 0.54 & 36.61 & 8.99  & 85.09 & 41.44  &  53.63 &  \cellcolor{cyan!20} 52.35 \\
\bottomrule
\end{tabular}%
}
\caption{Aggregated statistics across agreement classes for LiveNLI and VariErr, based on how many annotators agree on the NLI label. We report the label entropy, the percentages of support items, corresponding category agreement, and average pairwise explanation similarity.
Color coding highlights relative deviations within each dataset: cells shaded in \textcolor{cyan!60}{blue} or \textcolor{purple!60}{red} indicate values that are higher or lower than expected given the level of label agreement. 
Darker shades correspond to larger deviations in ranking, and lighter shades indicate smaller deviations.}
\label{tab:agreement_levels}
\end{table*}

\paragraph{Reasoning-Type Preferences}
Figure~\ref{fig:annotator_taxonomy} presents the distribution of explanation categories used by the annotators. Individual differences emerge in how annotators ground their inferences. 
For VariErr, annotator \texttt{0} shows a dominant reliance on \textit{Absence of Mention} (44.44\%) and \textit{Semantic} reasoning (24.31\%), with only minimal use of world knowledge-based categories such as \textit{Factual Knowledge} (0.69\%) and \textit{Inferential Knowledge} (13.66\%). This pattern suggests a preference for surface-level paraphrastic inference, rather than deeper reasoning. In contrast, Annotator \texttt{1} exhibits a more balanced distribution, with moderate use of the reasoning strategies. Annotator \texttt{2} stands out with a very strong emphasis on \textit{Inferential Knowledge} (30.29\%), while still relying on \textit{Semantic} explanations (24.48\%). This suggests a knowledge-intensive reasoning, grounded in world knowledge and inferencing beyond what is stated. Similarly, Annotator \texttt{3} relies more on knowledge-based reasoning, using \textit{Inferential Knowledge} (12.3\%) and \textit{Factual Knowledge} (3.2\%) more frequently.

For LiveNLI, several trends emerge from the distribution. First, \textit{Semantic} and \textit{Absence of Mention} explanations are consistently among the most frequently used categories across annotators, suggesting that both lexical-semantic inferences and missing information play a central role. Second, we observe notable variation in the use of \textit{Inferential Knowledge}: annotator \texttt{w4} relies on this category in over half of their explanations (54.25\%), while \texttt{w1} and \texttt{w8} use it far less frequently (20.00\% and 16.41\% respectively), indicating divergent preferences in relying on external world knowledge. Similarly, \textit{Syntactic} explanations are prominent for \texttt{w8} (21.88\%) compared to the others, reflecting a possible inclination toward structural reasoning. Conversely, \textit{Logical Conflict}, \textit{Factual Knowledge}, and \textit{Coreference} are relatively rare across annotators, suggesting these reasoning categories are less frequently invoked or salient in this LiveNLI subset.

Overall, the annotator-level analysis facilitated by \taxonomy{} over the two datasets reveals that \textbf{different annotators tend to adopt distinct reasoning strategies}---arriving at different NLI labels for the same premise–hypothesis pairs. Observing only the distribution of NLI labels is insufficient to uncover the underlying reasoning rationales. To gain a deeper understanding of annotator behavior, we next conduct a fine-grained item-level analysis that disentangles variation in reasoning from variation in label assignment.

\section{Measuring and Interpreting Agreement}\label{sec:per_item_analysis}

Building on observations in \S\ref{sec:label-category-distribution} that NLI label distributions are insufficient to uncover reasoning rationales of individual annotators, this section takes a closer look at how to measure and interpret agreement and disagreement in NLI tasks \textit{at the instance level}.
We first quantify annotator agreement across three dimensions: NLI labels, explanation categories, and textual similarity between explanations. We then compute pairwise agreement and visualize it using conditional probability heatmaps to examine the relationship between explanation categories and labeling.
 Finally, we present an example that is covered in both datasets to illustrate how annotators may align or diverge in their label decisions and explanations.

\subsection{Quantifying Annotator Agreement Beyond Labels}

To gain a clearer picture of annotator agreement on NLI labels and \taxonomy{} categories, we group NLI items based on the set of annotators who labeled them and how often these annotators agreed with each other.
We define four NLI \textbf{label agreement} classes:
\textit{Full Agreement} (4-0-0) indicates all four annotators assigned the same label;
\textit{Partial Agreement }(3-1-0) refers to cases where three annotators agreed while one differed;
\textit{Two Pairs Agreement} (2-2-0) denotes a balanced split, with two annotators agreeing on one label and the other two on a different label;
and \textit{Divergent} (2-1-1) captures maximal disagreement, where three different labels are assigned.

\textbf{Category agreement} is measured via the average Jaccard similarity between the explanation categories. Since each explanation has only one category, this reduces to computing the proportion of explanation pairs that share the same category:
\begin{align}
\text{Jaccard}(a, b) = 
\begin{cases}
1, & \text{if } a = b \\
0, & \text{otherwise}
\end{cases}
\end{align}

For example, for an item with four annotators ($\binom{4}{2}=6$ pairs), the LiveNLI example in Table~\ref{tab:example} shows a 2--2 split across categories (2/6 matches), while the VariErr example shows a 3--1 split (3/6 matches), corresponding to agreement scores of $0.33$ and $0.5$.

We follow \citet{chen-etal-2025-rose} and \citet{hong2025litexlinguistictaxonomyexplanations} to quantify \textbf{textual similarities between explanations} using measures from \citet{giulianelli-etal-2023-comes}. 
Lexical and syntactic similarities evaluate overlapping unigrams and bigrams on tokens and POS tags.
We use cosine and Euclidean to measure semantic similarity between sentence embeddings, obtained using the \texttt{all-distilroberta-v1}\footnote{\url{https://huggingface.co/sentence-transformers/all-distilroberta-v1}} model from SentenceTransformers \cite{sanh2020distilbertdistilledversionbert}.
Scores are averaged pairwise across four explanations.

Table~\ref{tab:agreement_levels} summarizes the label entropy, number of supporting items, category agreement, and average explanation similarity for each label agreement class.
Examining across evaluation metrics, the color coding highlights deviations in ranking across agreement classes relative to label agreement, though the absolute differences remain small.

VariErr generally exhibits less ranking deviation than LiveNLI, with a matched ranking between label agreement class and category agreement, and only light deviations in textual similarity measures. 
Full agreement is also considerably more frequent in VariErr (43.75\%) than in LiveNLI (21.74\%), indicating that annotators tend to reach agreement more easily in VariErr.
Moreover, all ranking deviations in VariErr concern an additional NLI label: between full and partial (4-0-0 vs.\ 3-\textbf{1}-0) and between two pairs and divergent (2-2-0 vs.\ 2-1-\textbf{1}), whereas many LiveNLI ranking deviations stem from different distributions of the same labels, i.e., between partial and two pairs (\textbf{3}-\textbf{1}-0 vs.\ \textbf{2}-\textbf{2}-0). 

Looking at the text similarity measures, we found that cosine similarity resonates the most with label agreement on both datasets, exhibiting moderate differences in scores across classes and only minor deviations in ranking on LiveNLI between partial and two pairs. 
More interestingly, the pattern of cosine similarity aligns more closely with category agreement than label agreement. 
This observation \textbf{tentatively suggests that shared reasoning categories may better capture the semantic similarity of explanations than label agreement}.

\subsection{Pairwise Annotator Agreement on Reasoning Category and Labeling}

\begin{figure}[t]
    \centering
    \includegraphics[width=\columnwidth]{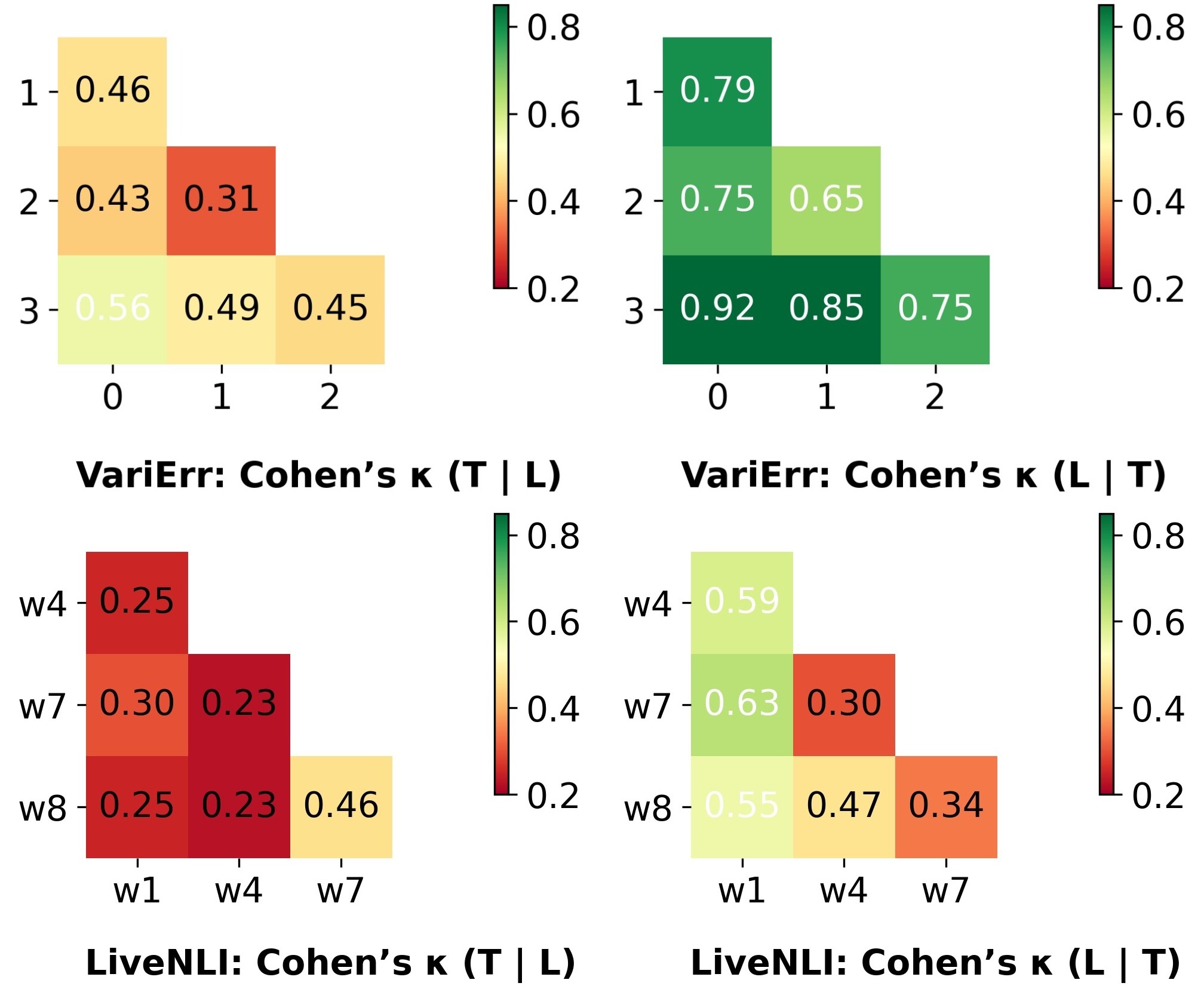}
    \caption{Pairwise annotator agreement (conditional Cohen's $\kappa$) between taxonomy matches (T) and label matches (L).}
    \label{fig:heatmap}
\end{figure}

\begin{table*}[t]
\centering
\renewcommand{\arraystretch}{1.2}
\setlength{\tabcolsep}{2.7pt}
\small 

\begin{tabular}{p{1.5cm}|C{0.7cm}|p{1.7cm}|p{8.2cm}|C{2.9cm}}
\toprule

\multicolumn{5}{p{\dimexpr\textwidth-2\tabcolsep-2\arrayrulewidth\relax}}{
\textbf{Premise:} The author began with a set of hunches or hypotheses about what can go wrong in agency management, and what would be evidence supporting—or contradicting—these hypotheses.
} \\

\multicolumn{5}{p{\dimexpr\textwidth-2\tabcolsep-2\arrayrulewidth\relax}}{
\textbf{Hypothesis:} The hunches provided by the author weren't realistic as it pertains to agency management.
} \\

\multicolumn{5}{p{\dimexpr\textwidth-2\tabcolsep-2\arrayrulewidth\relax}}{
[CHAOSNLI] [E,N,C]: [0.64, 0.06, 0.30]
} \\

\midrule
\textbf{Data} & \textbf{Ann.} & \makecell{\textbf{NLI}\\\textbf{Label}} & \textbf{Explanation} & \makecell{\textbf{Exp.}\\\textbf{Category}} \\
\midrule

\multirow{4}{*}{LiveNLI}
& \texttt{w1} & Neutral
& \makecell[tl]{The context notes that the hunches were provided,but there is no \\ information on their veracity or plausibility. Thus, the statement \\could be true or false, as it is not known whether they were \\realistic based on the provided information.}
& \makecell[tl]{Absence of Mention} \\

& \texttt{w4} & Contradiction
& \makecell[tl]{If an author is planning to write about a certain topic, they likely \\have enough knowledge to form informed opinions. Thus, it is \\much more likely that the statement is false, since these opinions \\would be at least somewhat realistic.}
& \makecell[tl]{Inferential Knowledge} \\

& \texttt{w7} & Contradiction
& \makecell[tl]{The author gave evidence to support the hunches,
it is unlikely\\ that the hunches were unrealistic.}
& \makecell[tl]{Inferential Knowledge} \\

& \texttt{w8} & Contradiction
& \makecell[tl]{It was not stated that the hunches were unrealistic.}
& \makecell[tl]{Absence of Mention} \\

\midrule

\multirow{4}{*}{VariErr}
& \texttt{0} & Neutral
& \makecell[tl]{It is not clear whether the hunches provided by the author were \\realistic or not.}
& \makecell[tl]{Absence of Mention} \\

& \texttt{1} & Neutral
& \makecell[tl]{It is not clear how realistic the hypotheses were.}
& \makecell[tl]{Absence of Mention} \\

& \texttt{2} & Neutral
& \makecell[tl]{The judgment of the hunches is not given in the context.}
& \makecell[tl]{Absence of Mention} \\

& \texttt{3} & Contradiction
& \makecell[tl]{The hunches could be realistic, as author provides potential \\evidence supporting these hypotheses.}
& \makecell[tl]{Inferential Knowledge} \\

\bottomrule
\end{tabular}

\caption{An NLI item annotated in both LiveNLI and VariErr. Despite identical inputs, annotators diverge in NLI labels and explanation strategies.}
\label{tab:example}

\end{table*}

To quantify how annotators align explanation strategies with label assignments, we examine pairwise agreement patterns between explanation categories (reasoning) and NLI labels (labeling) using conditional Cohen's $\kappa$ scores (Figure~\ref{fig:heatmap}). 
This analysis focuses on the interaction between explanation categories and NLI labels, allowing us to examine variation across both.
Specifically, for each pair of annotators, we compute two conditional $\kappa$ scores:
\begin{equation}
\kappa(T \mid L),
\label{eq:kappa-t-given-l}
\end{equation}
which measures taxonomy category agreement (reasoning alignment) on the subset of instances where the two annotators agree on the NLI label, and
\begin{equation}
\kappa(L \mid T).
\label{eq:kappa-l-given-t}
\end{equation}
which measures label agreement on the subset of instances where the two annotators match in their reasoning categories.
Unlike raw agreement, Cohen's $\kappa$ adjusts for chance, providing a more robust 
measure of inter-annotator alignment.

VariErr shows higher agreement than LiveNLI under both conditional $\kappa$ metric, indicating more stable reasoning–label mappings in VariErr, whereas LiveNLI reflects greater variation in annotator interpretations.
Across both datasets, annotators show higher $\kappa$ for labels conditioned on taxonomy category matches \(\kappa(L \mid T)\) than for taxonomy category conditioned on label matches \(\kappa(T \mid L)\). Our approach can explain the asymmetry between the two conditional probabilities. 
When annotators share the same taxonomy categories, they are highly likely to assign the same label, indicating that divergence in NLI label assignments exists but occurs less frequently. In contrast, when annotators agree on the final label, their reasoning categories often diverge, suggesting that \textbf{divergence in the explanation reasoning categories is relatively more common than that in the labeling}.
Therefore, to capture label variation more accurately and informatively, it is crucial to focus on the reasoning explicit expressed through free-text explanations, as it is the dominant source of variation.

\subsection{Example: Diverging Interpretations on the Same NLI Instance}\label{sec:case-study}

To conclude our analysis, we present an example that illustrates how annotators from two different datasets interpret the same NLI instance in divergent ways, both in the reasons exhibited in their explanations and in the label they choose. 
Table~\ref{tab:example} shows one item annotated by the eight annotators we analyzed in the earlier sections, four from LiveNLI and four from VariErr. For additional qualitative illustrations, further examples are provided in the Appendix~\ref{sec:overlap_examples}.

Comparing the two datasets, we observe that while the sets of chosen NLI labels (neutral and contradiction) and explanation categories (\textit{Absence of Mention} and \textit{Inferential Knowledge}) are the same, the distribution of these choices differs: in LiveNLI, three annotators opted for contradiction, whereas in VariErr, three chose neutral. In terms of explanation categories, LiveNLI annotators are evenly split, with two selecting \textit{Absence of Mention} and two choosing \textit{Inferential Knowledge}. In VariErr, three annotators attribute their reasoning to \textit{Absence of Mention}, while one goes for \textit{Inferential Knowledge}.

The example further illustrates patterns of agreement and disagreement in NLI annotations. Among LiveNLI annotators, \texttt{w4}, \texttt{w7}, and \texttt{w8} agree on the NLI label (contradiction), but provide explanations grounded in different reasoning categories, namely \textit{Inferential Knowledge} vs.\ \textit{Absence of Mention}, showcasing within-label variation and divergence at reasoning. In contrast, \texttt{w1} and \texttt{w8} both provide explanations categorized as \textit{Absence of Mention}, yet arrive at different NLI labels (neutral vs.\ contradiction), pointing to divergence at the labeling.
Meanwhile, VariErr annotators demonstrate both label and reasoning agreement. Annotators \texttt{0}, \texttt{1}, and \texttt{2} all classify the instance as neutral, supported by nearly identical explanations and shared categorization as \textit{Absence of Mention}. This coherence suggests a degree of alignment between reasoning categories and NLI labels, from interpretive rationale to label decision. Annotator \texttt{3} stands out with Inferential Knowledge reasoning and a contradiction label, further emphasizing how divergent reasoning can lead to different label choices.

Overall, this example shows \textbf{how combining NLI labels with explanation categories reveals deeper patterns of disagreement and agreement}---distinctions that would remain hidden if one only considered label distributions alone, while reasoning information further illuminates the underlying sources of variation.

\section{Conclusion}
Understanding why annotators diverge is key to interpreting NLI labels. We extend \taxonomy{} to two NLI datasets with free-text explanations and jointly analyze within-label and cross-annotator label variation. By combining labels, taxonomy categories, and explanation texts, we uncover reasoning patterns that label distributions do not capture. Across analyses, taxonomy categories track explanation-text similarity more closely than labels, emphasizing reasoning paths over surface label agreement.

Our results suggest three broader implications.
First, NLI explanations differ in what they ground the decision on, and these differences correlate with how often annotators agree or disagree.
Second, agreement in reasoning type is more predictive of label agreement than shared labels are of reasoning-type agreement, indicating that reasoning provides a key trace of variation.
Third, annotators exhibit distinct reasoning profiles, which can lead to divergent labels on the same instances.
These findings motivate future NLI work to explicitly track reasoning categories during dataset construction (e.g., which categories appear and which ones tend to trigger disagreement) and to use reasoning categories to stratify evaluation sets, offering a more diagnostic view of where models succeed or fail.

More broadly, labels and even aggregated label distributions can hide how annotators arrive at their interpretations.
Complementing labels with explanations provides a clearer window into patterns of agreement and disagreement among annotators.
While the ideal explanation format remains open, free-text explanations retain a unique advantage in revealing how different interpretations can emerge from the same input.

Future work can extend \taxonomy{} to allow multiple reasoning strategies per explanation, capturing more interactions.
Modeling annotator backgrounds may further reveal systematic sources of disagreement.
Finally, integrating this framework with explanation generation could improve the quality and evaluation of model rationales, and applying it to other domains would test its generality. Annotations and analysis are publicly available at \href{https://github.com/mainlp/LiTEx-NLI-extension}{https://github.com/mainlp/LiTEx-NLI-extension}.

\section*{Limitations}
Our work has several limitations.
First, our analysis relies on the category set defined in \taxonomy{}, which may not fully capture the complexity or compositionality of human reasoning; extending the taxonomy to allow multiple categories per explanation is an important direction.

Second, our datasets are substantially smaller than large-scale NLI benchmarks such as SNLI or MNLI.
However, the analysis is based on 1{,}404 explanations from LiveNLI and 1{,}933 explanations from VariErr, which is relatively large for manually categorized free-text explanations.
Accordingly, our goal is not to estimate fine-grained population statistics for all possible annotators, but to uncover robust qualitative and quantitative patterns in how reasoning categories relate to label variation.

Third, our annotator-level analysis focuses on annotators with sufficient per-annotator coverage: we analyze four annotators who each contributed explanations for over 100 items on a shared set of instances, providing within-annotator evidence for comparing reasoning profiles.
Nevertheless, including more annotators and more diverse annotator populations would strengthen our work.

Finally, we do not directly model annotator-pool effects (e.g., background knowledge, fatigue, or instruction framing), and our similarity measures (e.g., sentence-embedding cosine similarity and Jaccard agreement over categories) provide only a partial view of reasoning variation, potentially missing deeper pragmatic or structural divergences.

\section*{Ethical considerations} 
We do not foresee any ethical concerns associated with this work. All analyses were conducted using publicly available datasets and models. No private or sensitive information was used. 

\section*{Acknowledgments}

We would like to thank the anonymous reviewers for their insightful comments and valuable suggestions. We are also grateful to the members of the MaiNLP Lab and the Natural Language Processing (NLP) Working Group at the University of Vienna for their constructive feedback on earlier drafts of this paper. In particular, we sincerely appreciate the helpful input from Verena Blaschke, Andreas Säuberli, and Yang Janet Liu.

This research was supported by the Vienna Science and Technology Fund (WWTF) under grant [10.47379/VRG19008], Knowledge-Infused Deep Learning for Natural Language Processing. Beiduo Chen acknowledges support from the European Laboratory for Learning and Intelligent Systems (ELLIS) PhD program. Marie-Catherine de Marneffe is a Research Associate of the Fonds de la Recherche Scientifique – FNRS. This work was also supported by the ERC Consolidator Grant DIALECT (101043235).

\paragraph{Use of AI Assistants}
The authors acknowledge the use of ChatGPT for
correcting grammatical errors and obtaining suggestions to enhance the coherence of the initial manuscript.

\appendix

\section{Label Distribution per NLI Item in LiveNLI and VariErr}\label{sec:label_distribution}

To better understand the distribution characteristics of NLI labels in the selected two datasets, we visualize the aggregated label probabilities for each item. 

\begin{figure}[h]
  \includegraphics[width=\columnwidth]{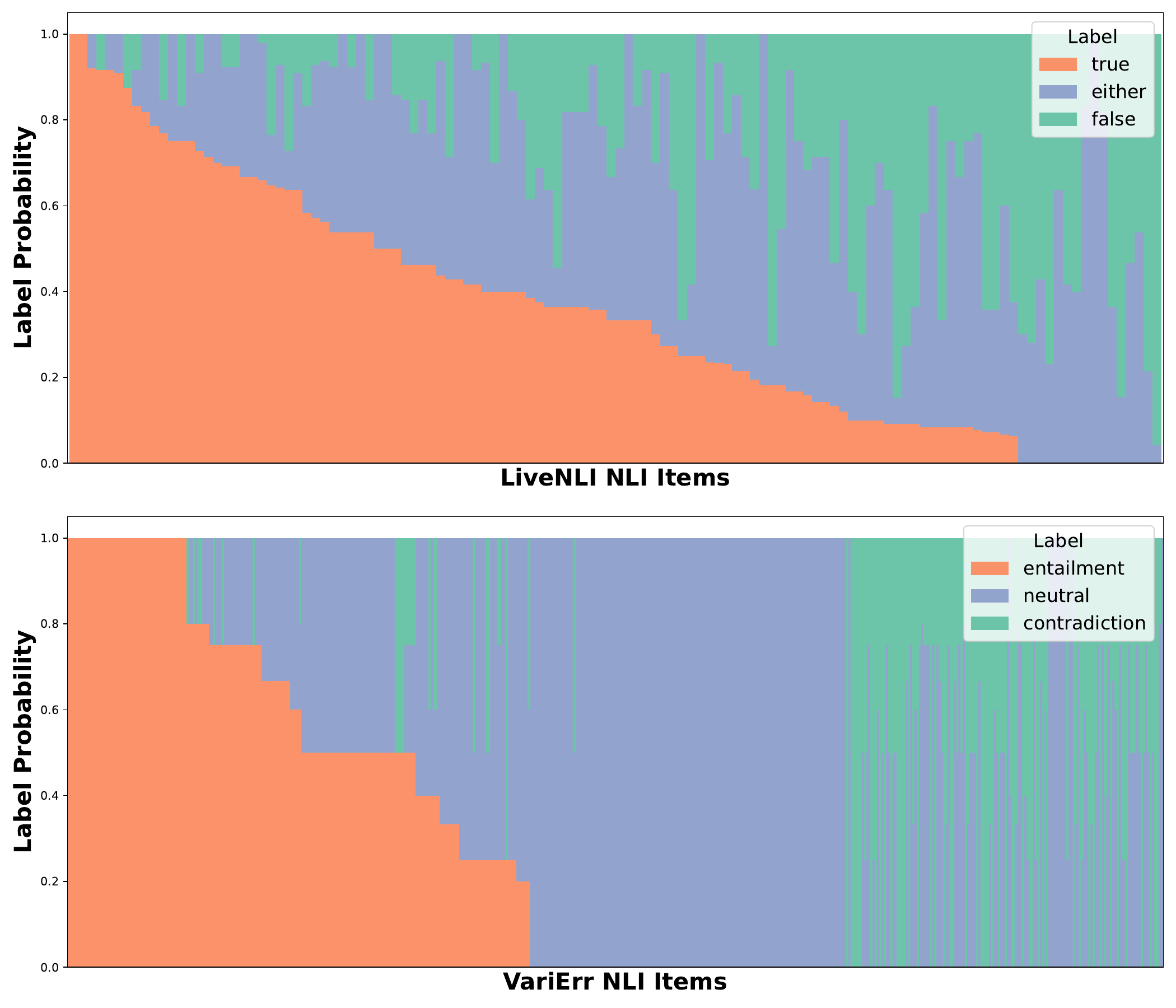}
  \caption{Normalized label distribution per NLI item in LiveNLI and VariErr. Items are sorted by the proportion of \textit{entailment}/\textit{true} labels.}
  \label{fig:nli-label-distribution}
\end{figure}

Figure~\ref{fig:nli-label-distribution} presents stacked bar charts of the normalized NLI label distributions across all items, sorted by the proportion of \textit{entailment}/\textit{true} labels to highlight overall patterns of annotator agreement. Both datasets reveal a range of label variation, with many items reflecting ambiguity or disagreement among annotators. However, their distributions differ in characteristic ways. LiveNLI shows greater diversity, particularly in cases where the \textit{either} label dominates or co-occurs substantially with the other two categories. In contrast, VariErr exhibits more concentrated distributions, with fewer items displaying high levels of ambiguity. Nonetheless, some items still reveal substantial variation in label assignment, pointing to challenging or underspecified NLI cases. These findings motivate our subsequent analysis of how explanation similarity varies across explanation categories and NLI labels.

\section{Inter-annotation agreement (IAA) results of LiveNLI and VariErr} \label{sec:IAA}

To further strengthen the validity and transparency of our annotation process, we report detailed IAA statistics for both \textbf{LiveNLI} and \textbf{VariErr}.

As shown in Table~\ref{tab:iaa-category-distribution}, we observe that the distribution of explanation categories is notably non-uniform across both datasets. Certain categories, such as \textit{Factual Knowledge}, occur only rarely compared to more dominant categories like \textit{Absence of Mention} or \textit{Pragmatic}. This skewed distribution is consistent with observations reported in the original LITEX framework \citep{hong2025litexlinguistictaxonomyexplanations}. Given this imbalance, our analysis does not aim to draw fine-grained conclusions about every individual explanation category.

\begin{table}[h]
\centering
\small
\resizebox{\linewidth}{!}{%
\begin{tabular}{lcccc}
\toprule
 & \multicolumn{2}{c}{\textbf{LiveNLI}} & \multicolumn{2}{c}{\textbf{VariErr}} \\
\cmidrule(lr){2-3} \cmidrule(lr){4-5}
\textbf{Category} & \textbf{Annotator 1} & \textbf{Annotator 2} & \textbf{Annotator 1} & \textbf{Annotator 2} \\
\midrule
Absence of Mention & 39 & 41 & 36 & 38 \\
Pragmatic & 21 & 14 & 19 & 17 \\
Inferential Knowledge & 18 & 14 & 16 & 15 \\
Semantic & 6 & 16 & 10 & 12 \\
Logic Conflict & 6 & 5 & 7 & 6 \\
Syntactic & 5 & 4 & 5 & 4 \\
Factual Knowledge & 3 & 4 & 4 & 5 \\
Coreference & 2 & 2 & 3 & 3 \\
\midrule
\textbf{Total} & 100 & 100 & 100 & 100 \\
\bottomrule
\end{tabular}
}
\caption{Distribution of \taxonomy{} categories in the 100-item IAA samples for LiveNLI and VariErr.}
\label{tab:iaa-category-distribution}
\end{table}

Table~\ref{tab:iaa-agreement} reports per-category agreement across the two datasets. We observe systematic variation in IAA: high-frequency categories such as \textit{Absence of Mention} show consistently high agreement (90\% and 88\%), indicating that they are salient and reliably identifiable. Categories like \textit{Inferential Knowledge} and \textit{Logic Conflict} also achieve relatively strong agreement (around 80\%), reflecting well-defined reasoning patterns. In contrast, \textit{Factual Knowledge} shows lower agreement (67\% and 70\%), likely due to limited sample size and overlap with categories such as \textit{Inferential Knowledge}.

\begin{table}[h]
\centering
\footnotesize
\resizebox{\linewidth}{!}{%
\begin{tabular}{lccccc}
\toprule
 & \multicolumn{2}{c}{\textbf{LiveNLI}} & \multicolumn{2}{c}{\textbf{VariErr}} \\
\cmidrule(lr){2-3} \cmidrule(lr){4-5}
\textbf{Category} & \textbf{Support} & \textbf{Agree (\%)} & \textbf{Support} & \textbf{Agree (\%)} \\
\midrule
Absence of Mention & 40 & 90 & 38 & 88 \\
Pragmatic & 17 & 76 & 18 & 78 \\
Inferential Knowledge & 16 & 81 & 15 & 80 \\
Semantic & 11 & 73 & 12 & 75 \\
Logic Conflict & 6 & 83 & 7 & 82 \\
Syntactic & 4 & 75 & 5 & 74 \\
Factual Knowledge & 3 & 67 & 4 & 70 \\
Coreference & 2 & 100 & 3 & 100 \\
\bottomrule
\end{tabular}
}
\caption{Per-category raw agreement for LiveNLI and VariErr.}
\label{tab:iaa-agreement}
\end{table}

\section{Annotator-wise distribution of NLI labels and explanation categories} \label{sec:distribution_counts}

Table~\ref{tab:annotator_tracking} presents the detailed NLI label distributions and explanation category preferences for the four annotators in LiveNLI and VariErr, computed over the intersection of items they jointly annotated in each dataset. This table provides the full statistics underlying the summary reported in Section~\ref{sec:label-category-distribution}.

\begin{table*}[t]
\centering
\small
\setlength{\tabcolsep}{5pt}
\begin{tabular}{c|r|ccc|cccccccc}
\toprule
\textbf{Annotator ID} & \textbf{\#Ex} & \textbf{Ent} & \textbf{Neu} & \textbf{Con} & \textbf{Coref} & \textbf{Synt} & \textbf{Sem} & \textbf{Prag} & \textbf{Abs} & \textbf{Logic} & \textbf{FK} & \textbf{IK} \\
\midrule
\rowcolor{mycolor}\multicolumn{13}{l}{\textit{VariErr}\rule{0pt}{8pt}} \\
\texttt{0} & 432 & 29.40 & 51.62 & 18.98 & 1.85 & 2.31 & 24.31 & 3.70 & 44.44 & 9.03 & 0.69 & 13.66 \\
\texttt{1} & 446 & 30.04 & 55.83 & 14.13 & 0.90 & 10.09 & 19.06 & 4.48 & 37.67 & 8.30 & 1.79 & 17.71 \\
\texttt{2} & 482 & 21.37 & 59.13 & 19.50 & 1.04 & 2.70 & 24.48 & 1.87 & 28.22 & 8.92 & 2.49 & 30.29 \\
\texttt{3} & 439 & 30.75 & 44.42 & 24.83 & 0.68 & 5.92 & 28.93 & 3.64 & 32.80 & 12.53 & 3.19 & 12.30 \\
\rowcolor{mycolor}\multicolumn{13}{l}{\textit{LiveNLI}\rule{0pt}{8pt}} \\
\texttt{w1} & 140 & 39.29 & 34.29 & 25.71 & 1.43 & 4.29 & 17.86 & 17.14 & 27.14 & 12.14 & 0.00 & 20.00 \\
\texttt{w4} & 153 & 39.87 & 26.14 & 33.99 & 2.61 & 4.58 & 13.07 & 11.76 & 9.15 & 3.27 & 1.31 & 54.25 \\
\texttt{w7} & 144 & 27.78 & 52.08 & 20.14 & 1.39 & 6.25 & 15.97 & 15.97 & 25.69 & 5.56 & 2.78 & 26.39 \\
\texttt{w8} & 128 & 41.41 & 28.13 & 30.47 & 1.56 & 21.88 & 18.75 & 10.16 & 22.66 & 5.47 & 3.13 & 16.41 \\
\bottomrule
\end{tabular}
\caption{Annotator-wise distribution of NLI labels (entailment, neutral, contradiction) and explanation categories in LiveNLI and VariErr. Percentages (\%) are shown for both types of distributions.}
\label{tab:annotator_tracking}
\end{table*}

\section{NLI items annotated in both LiveNLI and VariErr}
\label{sec:overlap_examples}

Table~\ref{tab:examples} presents more illustrative examples of NLI items that were annotated in both LiveNLI and VariErr by the selected annotator groups. These examples provide additional qualitative context for the agreement patterns discussed in the main text.

\begin{table*}[t]
\centering
\renewcommand{\arraystretch}{1.2}
\setlength{\tabcolsep}{5pt}
\resizebox{\textwidth}{!}{%
\begin{tabular}{l|l|l|p{19cm}|l}
\toprule
\multicolumn{5}{l}{
    \textbf{Premise:} If you have any questions about this report, please contact Henry R. Wray, Senior Associate General Counsel, at (202) 512-8581.
} \\
\multicolumn{5}{l}{
    \textbf{Hypothesis:} Henry R. Wray can be reached at (555) 512-8581.
} \\
\multicolumn{5}{l}{
    [CHAOSNLI] [E,N,C]: [0.21, 0.76, 0.03]
} 
\\
\midrule
\textbf{Dataset} & \textbf{Annotator} & \textbf{NLI Label} & \textbf{Explanation} & \textbf{Explanation Category} \\
\midrule
\multirow{4}{*}{LiveNLI} 
& \texttt{w1} & Contradiction & The two phone numbers provided for Wray do not match, thus the statement is likely to be false since it does not present the same number as in the context. & Logic Conflict \\
& \texttt{w4} & Contradiction & Henry R. Wray's actual phone number is listed in the context.  While it's true he could have multiple contact numbers, it's more likely that the context is his only contact number since it's connected to his business. & Inferential Knowledge \\
& \texttt{w7} & Contradiction & The phone number for Henry Wray is (202) 512-8581, so (555) 512-8581 is probably not going to reach him. & Inferential Knowledge \\
& \texttt{w8} & Contradiction & Henry R. Wray can be reached at a difference number which is (202) 512-8581. & Semantic \\
\midrule
\multirow{4}{*}{VariErr} 
& \texttt{0} & Contradiction & The number of Henry R. Wray mentioned in the statement is wrong. & Logic Conflict \\
& \texttt{1} & Contradiction & The phone number starts with (202) not with (555). & Logic Conflict \\
& \texttt{2} & Contradiction & The number is wrong, it should be (202) 512-8581 not (555) 512-8581. & Logic Conflict \\
& \texttt{3} & Contradiction & wrong phone numbers. & Logic Conflict \\
\midrule
\multicolumn{5}{l}{
    \textbf{Premise:} It is, as you see, highly magnified.
} \\
\multicolumn{5}{l}{
    \textbf{Hypothesis:} It is plain for you to see that it is amplified.
} \\
\multicolumn{5}{l}{
    [CHAOSNLI] [E,N,C]: [0.76, 0.10, 0.14]
} 
\\
\midrule
\multirow{4}{*}{LiveNLI} 
& \texttt{w1} & Entailment & Magnification is a form of amplification.  Therefore, if something is easily seen as magnified, it follows that it's plain to see it's amplified. & Inferential Knowledge \\
& \texttt{w4} & Entailment & Whatever the speaker is pointing out is clearly magnified or "amplified", and the person they are showing this to is merely having this pointed out to them, making the statement most likely true. & Inferential Knowledge \\
& \texttt{w7} & Entailment & The words magnified and amplified have the same basic meaning, so the statement agrees with the context. & Semantic \\
& \texttt{w8} & Entailment & Both amplified and magnified mean the same thing so the statement is true. & Semantic \\
\midrule
\multirow{4}{*}{VariErr} 
& \texttt{0} & Entailment & Both the context and the statement suggest that it is magnified. & Semantic \\
& \texttt{1} & Entailment & The statement is a paraphrase of the context. & Syntactic \\
& \texttt{2} & Entailment & \"highly magnified\" can be interpreted \"amplified\". & Semantic \\
& \texttt{3} & Entailment & It can be seen, and it is magnified. & Semantic \\
\midrule
\multicolumn{5}{l}{
    \textbf{Premise:} A clean, wholesome-looking woman opened it.
} \\
\multicolumn{5}{l}{
    \textbf{Hypothesis:} The woman was trying to be desecrate.
} \\
\multicolumn{5}{l}{
    [CHAOSNLI] [E,N,C]: [0.68, 0.31, 0.01]
} 
\\
\midrule
\multirow{4}{*}{LiveNLI} 
& \texttt{w1} & Contradiction & The context notes that the woman is clean and wholesome-looking while the statement notes that the woman was being disrespectful, which is not compatible. Thus, it is likely to be false. & Semantic \\
& \texttt{w4} & Neutral & The statement is nonsensical.  Hence there's no information in it, either true or false, to be compared to the context. & Absence of Mention \\
& \texttt{w7} & Neutral & Just because the woman was wholesome-looking does not mean that she was acting in a discreet manner. & Inferential Knowledge \\
& \texttt{w8} & Contradiction & The woman was described as wholesome and wouldn't desecrate something. & Semantic \\
\midrule
\multirow{4}{*}{VariErr} 
& \texttt{0} & Neutral & The context doesn't mention anything about desecration. & Absence of Mention \\
& \texttt{1} & Neutral & It's not clear what the woman was trying to be. & Absence of Mention \\
& \texttt{2} & Neutral & The attempt of the woman is not given in the context. & Absence of Mention \\
& \texttt{3} & Contradiction & Context is a compliment, statement is a negative comment. & Logic Conflict \\
\bottomrule
\end{tabular}
}
\caption{Examples of NLI items annotated in both LiveNLI and VariErr by the selected annotator groups (w1, w4, w7, w8 for LiveNLI; 0, 1, 2, 3 for VariErr).}
\label{tab:examples}
\end{table*}


\begin{thebibliography}{25}
\providecommand{\natexlab}[1]{#1}

\bibitem[{Bowman et~al.(2015)Bowman, Angeli, Potts, and Manning}]{bowman-etal-2015-large}
Samuel~R. Bowman, Gabor Angeli, Christopher Potts, and Christopher~D. Manning. 2015.
\newblock \href {https://doi.org/10.18653/v1/D15-1075} {A {L}arge {A}nnotated {C}orpus for {L}earning {N}atural {L}anguage {I}nference}.
\newblock In \emph{Proceedings of the 2015 Conference on Empirical Methods in Natural Language Processing}, pages 632--642, Lisbon, Portugal. Association for Computational Linguistics.

\bibitem[{Camburu et~al.(2018)Camburu, Rockt\"{a}schel, Lukasiewicz, and Blunsom}]{camburu2018esnlinaturallanguageinference}
Oana-Maria Camburu, Tim Rockt\"{a}schel, Thomas Lukasiewicz, and Phil Blunsom. 2018.
\newblock \href {https://proceedings.neurips.cc/paper_files/paper/2018/file/4c7a167bb329bd92580a99ce422d6fa6-Paper.pdf} {e-{SNLI}: {N}atural {L}anguage {I}nference with {N}atural {L}anguage {E}xplanations}.
\newblock In \emph{Advances in Neural Information Processing Systems}, volume~31. Curran Associates, Inc.

\bibitem[{Chen et~al.(2025{\natexlab{a}})Chen, Liu, Korhonen, and Plank}]{chen-etal-2025-threading}
Beiduo Chen, Yang~Janet Liu, Anna Korhonen, and Barbara Plank. 2025{\natexlab{a}}.
\newblock \href {https://doi.org/10.18653/v1/2025.emnlp-main.1682} {{T}hreading the {N}eedle: {R}eweaving {C}hain-of-{T}hought {R}easoning to {E}xplain {H}uman {L}abel {V}ariation}.
\newblock In \emph{Proceedings of the 2025 Conference on Empirical Methods in Natural Language Processing}, pages 33111--33135, Suzhou, China. Association for Computational Linguistics.

\bibitem[{Chen et~al.(2025{\natexlab{b}})Chen, Peng, Korhonen, and Plank}]{chen-etal-2025-rose}
Beiduo Chen, Siyao Peng, Anna Korhonen, and Barbara Plank. 2025{\natexlab{b}}.
\newblock \href {https://doi.org/10.18653/v1/2025.findings-acl.562} {A {R}ose by {A}ny {O}ther {N}ame: {LLM}-{G}enerated {E}xplanations {Are} {G}ood {P}roxies for {H}uman {E}xplanations to {C}ollect {L}abel {D}istributions on {NLI}}.
\newblock In \emph{Findings of the Association for Computational Linguistics: ACL 2025}, pages 10777--10802, Vienna, Austria. Association for Computational Linguistics.

\bibitem[{Condoravdi et~al.(2003)Condoravdi, Crouch, de~Paiva, Stolle, and Bobrow}]{condoravdi-etal-2003-entailment}
Cleo Condoravdi, Dick Crouch, Valeria de~Paiva, Reinhard Stolle, and Daniel~G. Bobrow. 2003.
\newblock \href {https://aclanthology.org/W03-0906/} {Entailment, {I}ntensionality and {T}ext {U}nderstanding}.
\newblock In \emph{Proceedings of the {HLT}-{NAACL} 2003 Workshop on Text Meaning}, pages 38--45.

\bibitem[{Dagan et~al.(2005)Dagan, Glickman, and Magnini}]{inproceedings}
Ido Dagan, Oren Glickman, and Bernardo Magnini. 2005.
\newblock \href {https://doi.org/10.1007/11736790_9} {The {PASCAL} {R}ecognising {T}extual {E}ntailment {C}hallenge}.
\newblock In \emph{Machine Learning Challenges Workshop}, pages 177--190.

\bibitem[{de~Marneffe et~al.(2012)de~Marneffe, Manning, and Potts}]{demarneffe-etal-2012}
Marie-Catherine de~Marneffe, Christopher~D. Manning, and Christopher Potts. 2012.
\newblock \href {https://doi.org/10.1162/COLI_a_00097} {{D}id {I}t {H}appen? {T}he {P}ragmatic {C}omplexity of {V}eridicality {A}ssessment}.
\newblock \emph{Computational Linguistics}, 38(2):301--333.

\bibitem[{Giulianelli et~al.(2023)Giulianelli, Baan, Aziz, Fern{\'a}ndez, and Plank}]{giulianelli-etal-2023-comes}
Mario Giulianelli, Joris Baan, Wilker Aziz, Raquel Fern{\'a}ndez, and Barbara Plank. 2023.
\newblock \href {https://doi.org/10.18653/v1/2023.emnlp-main.887} {{W}hat {C}omes {N}ext? {E}valuating {U}ncertainty in {N}eural {T}ext {G}enerators {A}gainst {H}uman {P}roduction {V}ariability}.
\newblock In \emph{Proceedings of the 2023 Conference on Empirical Methods in Natural Language Processing}, pages 14349--14371, Singapore. Association for Computational Linguistics.

\bibitem[{Gubelmann et~al.(2023)Gubelmann, Kalouli, Niklaus, and Handschuh}]{gubelmann-etal-2023-truth}
Reto Gubelmann, Aikaterini-lida Kalouli, Christina Niklaus, and Siegfried Handschuh. 2023.
\newblock \href {https://doi.org/10.18653/v1/2023.starsem-1.4} {When {T}ruth {M}atters - {A}ddressing {P}ragmatic {C}ategories in {N}atural {L}anguage {I}nference ({NLI}) by {L}arge {L}anguage {M}odels ({LLM}s)}.
\newblock In \emph{Proceedings of the 12th Joint Conference on Lexical and Computational Semantics (*SEM 2023)}, pages 24--39, Toronto, Canada. Association for Computational Linguistics.

\bibitem[{Hong et~al.(2025)Hong, Chen, Peng, de~Marneffe, and Plank}]{hong2025litexlinguistictaxonomyexplanations}
Pingjun Hong, Beiduo Chen, Siyao Peng, Marie-Catherine de~Marneffe, and Barbara Plank. 2025.
\newblock \href {https://doi.org/10.18653/v1/2025.emnlp-main.1728} {{L}i{TE}x: {A} {L}inguistic {T}axonomy of {E}xplanations for {U}nderstanding {W}ithin-{L}abel {V}ariation in {N}atural {L}anguage {I}nference}.
\newblock In \emph{Proceedings of the 2025 Conference on Empirical Methods in Natural Language Processing}, pages 34065--34085, Suzhou, China. Association for Computational Linguistics.

\bibitem[{Jiang and de~Marneffe(2022)}]{jiang-marneffe-2022-investigating}
Nan-Jiang Jiang and Marie-Catherine de~Marneffe. 2022.
\newblock \href {https://doi.org/10.1162/tacl_a_00523} {{I}nvestigating {R}easons for {D}isagreement in {N}atural {L}anguage {I}nference}.
\newblock \emph{Transactions of the Association for Computational Linguistics}, 10:1357--1374.

\bibitem[{Jiang et~al.(2023)Jiang, Tan, and de~Marneffe}]{jiang-etal-2023-ecologically}
Nan-Jiang Jiang, Chenhao Tan, and Marie-Catherine de~Marneffe. 2023.
\newblock \href {https://doi.org/10.18653/v1/2023.findings-emnlp.712} {Ecologically {V}alid {E}xplanations for {L}abel {V}ariation in {NLI}}.
\newblock In \emph{Findings of the Association for Computational Linguistics: EMNLP 2023}, pages 10622--10633, Singapore. Association for Computational Linguistics.

\bibitem[{Kalouli et~al.(2023)Kalouli, Hu, Webb, Moss, and de~Paiva}]{kalouli-etal-2023-curing}
Aikaterini-Lida Kalouli, Hai Hu, Alexander~F. Webb, Lawrence~S. Moss, and Valeria de~Paiva. 2023.
\newblock \href {https://doi.org/10.1162/coli_a_00465} {Curing the {SICK} and {O}ther {NLI} {M}aladies}.
\newblock \emph{Computational Linguistics}, 49(1):199--243.

\bibitem[{Liu et~al.(2023)Liu, Wu, Michael, Suhr, West, Koller, Swayamdipta, Smith, and Choi}]{liu-etal-2023-afraid}
Alisa Liu, Zhaofeng Wu, Julian Michael, Alane Suhr, Peter West, Alexander Koller, Swabha Swayamdipta, Noah Smith, and Yejin Choi. 2023.
\newblock \href {https://doi.org/10.18653/v1/2023.emnlp-main.51} {We{'}re {A}fraid {L}anguage {M}odels {A}ren{'}t {M}odeling {A}mbiguity}.
\newblock In \emph{Proceedings of the 2023 Conference on Empirical Methods in Natural Language Processing}, pages 790--807, Singapore. Association for Computational Linguistics.

\bibitem[{Nie et~al.(2020{\natexlab{a}})Nie, Williams, Dinan, Bansal, Weston, and Kiela}]{nie-etal-2020-adversarial}
Yixin Nie, Adina Williams, Emily Dinan, Mohit Bansal, Jason Weston, and Douwe Kiela. 2020{\natexlab{a}}.
\newblock \href {https://doi.org/10.18653/v1/2020.acl-main.441} {Adversarial {NLI}: {A} {N}ew {B}enchmark for {N}atural {L}anguage {U}nderstanding}.
\newblock In \emph{Proceedings of the 58th Annual Meeting of the Association for Computational Linguistics}, pages 4885--4901, Online. Association for Computational Linguistics.

\bibitem[{Nie et~al.(2020{\natexlab{b}})Nie, Zhou, and Bansal}]{nie-etal-2020-learn}
Yixin Nie, Xiang Zhou, and Mohit Bansal. 2020{\natexlab{b}}.
\newblock \href {https://doi.org/10.18653/v1/2020.emnlp-main.734} {What {C}an {W}e {L}earn from {C}ollective {H}uman {O}pinions on {N}atural {L}anguage {I}nference {D}ata?}
\newblock In \emph{Proceedings of the 2020 Conference on Empirical Methods in Natural Language Processing (EMNLP)}, pages 9131--9143, Online. Association for Computational Linguistics.

\bibitem[{Nighojkar et~al.(2023)Nighojkar, Laverghetta~Jr., and Licato}]{nighojkar-etal-2023-strong}
Animesh Nighojkar, Antonio Laverghetta~Jr., and John Licato. 2023.
\newblock \href {https://doi.org/10.18653/v1/2023.law-1.20} {No {S}trong {F}eelings {O}ne {W}ay or {A}nother: {R}e-operationalizing {N}eutrality in {N}atural {L}anguage {I}nference}.
\newblock In \emph{Proceedings of the 17th Linguistic Annotation Workshop (LAW-XVII)}, pages 199--210, Toronto, Canada. Association for Computational Linguistics.

\bibitem[{Pavlick and Kwiatkowski(2019)}]{pavlick-kwiatkowski-2019-inherent}
Ellie Pavlick and Tom Kwiatkowski. 2019.
\newblock \href {https://doi.org/10.1162/tacl_a_00293} {Inherent {D}isagreements in {H}uman {T}extual {I}nferences}.
\newblock \emph{Transactions of the Association for Computational Linguistics}, 7:677--694.

\bibitem[{Plank(2022)}]{plank-2022-problem}
Barbara Plank. 2022.
\newblock \href {https://doi.org/10.18653/v1/2022.emnlp-main.731} {The ``{P}roblem'' of {H}uman {L}abel {V}ariation: {O}n {G}round {T}ruth in {D}ata, {M}odeling and {E}valuation}.
\newblock In \emph{Proceedings of the 2022 Conference on Empirical Methods in Natural Language Processing}, pages 10671--10682, Abu Dhabi, United Arab Emirates. Association for Computational Linguistics.

\bibitem[{Sanh et~al.(2019)Sanh, Debut, Chaumond, and Wolf}]{sanh2020distilbertdistilledversionbert}
Victor Sanh, Lysandre Debut, Julien Chaumond, and Thomas Wolf. 2019.
\newblock \href {https://arxiv.org/abs/1910.01108} {Distilbert, a distilled version of {BERT:} smaller, faster, cheaper and lighter}.
\newblock \emph{CoRR}, abs/1910.01108.

\bibitem[{Sanyal et~al.(2024)Sanyal, Xiao, Liu, Wang, and Ren}]{sanyal-etal-2024-machines}
Soumya Sanyal, Tianyi Xiao, Jiacheng Liu, Wenya Wang, and Xiang Ren. 2024.
\newblock \href {https://doi.org/10.18653/v1/2024.findings-acl.618} {Are {M}achines {B}etter at {C}omplex {R}easoning? {U}nveiling {H}uman-{M}achine {I}nference {G}aps in {E}ntailment {V}erification}.
\newblock In \emph{Findings of the Association for Computational Linguistics: ACL 2024}, pages 10361--10386, Bangkok, Thailand. Association for Computational Linguistics.

\bibitem[{Tan(2022)}]{tan-2022-diversity}
Chenhao Tan. 2022.
\newblock \href {https://doi.org/10.18653/v1/2022.naacl-main.158} {On the {D}iversity and {L}imits of {H}uman {E}xplanations}.
\newblock In \emph{Proceedings of the 2022 Conference of the North American Chapter of the Association for Computational Linguistics: Human Language Technologies}, pages 2173--2188, Seattle, United States. Association for Computational Linguistics.

\bibitem[{Uma et~al.(2022)Uma, Almanea, and Poesio}]{Uma2022ScalingAD}
Alexandra Uma, Dina Almanea, and Massimo Poesio. 2022.
\newblock \href {https://api.semanticscholar.org/CorpusID:247845571} {Scaling and {D}isagreements: {B}ias, {N}oise, and {A}mbiguity}.
\newblock \emph{Frontiers in Artificial Intelligence}, 5.

\bibitem[{Weber-Genzel et~al.(2024)Weber-Genzel, Peng, de~Marneffe, and Plank}]{weber-genzel-etal-2024-varierr}
Leon Weber-Genzel, Siyao Peng, Marie-Catherine de~Marneffe, and Barbara Plank. 2024.
\newblock \href {https://doi.org/10.18653/v1/2024.acl-long.123} {{V}ari{E}rr {NLI}: {S}eparating {A}nnotation {E}rror from {H}uman {L}abel {V}ariation}.
\newblock In \emph{Proceedings of the 62nd Annual Meeting of the Association for Computational Linguistics (Volume 1: Long Papers)}, pages 2256--2269, Bangkok, Thailand. Association for Computational Linguistics.

\bibitem[{Williams et~al.(2018)Williams, Nangia, and Bowman}]{williams-etal-2018-broad}
Adina Williams, Nikita Nangia, and Samuel Bowman. 2018.
\newblock \href {https://doi.org/10.18653/v1/N18-1101} {A {B}road-{C}overage {C}hallenge {C}orpus for {S}entence {U}nderstanding through {I}nference}.
\newblock In \emph{Proceedings of the 2018 Conference of the North {A}merican Chapter of the Association for Computational Linguistics: Human Language Technologies, Volume 1 (Long Papers)}, pages 1112--1122, New Orleans, Louisiana. Association for Computational Linguistics.

\end{thebibliography}
\end{document}